\documentclass[10pt]{article} 
\usepackage[accepted]{rlc}

\usepackage{amssymb}            
\usepackage{mathtools}          
\usepackage{mathrsfs}           
\usepackage{amsmath}
\mathtoolsset{showonlyrefs}     
\usepackage{graphicx}           
\usepackage{subcaption}         
\usepackage[space]{grffile}     
\usepackage{url}                

\title{Zero-shot cross-modal transfer of Reinforcement Learning policies through a Global Workspace}

\author{Léopold Maytié  \\
    leopold.maytie@univ-tlse3.fr \\
    CerCo, CNRS UMR5549\\
    Artificial and Natural Intel-\\
    ligence Toulouse Institute,\\
    Université de Toulouse
    \And
    Benjamin Devillers \\
    benjamin.devillers@cnrs.fr \\
    CerCo, CNRS UMR5549 \\
    Artificial and Natural Intel-\\
    ligence Toulouse Institute,\\
    Université de Toulouse
    \And
    Alexandre Arnold \\
    alexandre.arnold@airbus.com \\
    Airbus AI Research
    \And
    Rufin VanRullen \\
    rufin.vanrullen@cnrs.fr \\
    CerCo, CNRS UMR5549 \\
    Artificial and Natural Intel-\\
    ligence Toulouse Institute,\\
    Université de Toulouse}

\begin{document}

\maketitle

\begin{abstract}
Humans perceive the world through multiple senses, enabling them to create a comprehensive representation of their surroundings and to generalize information across domains. For instance, when a textual description of a scene is given, humans can mentally visualize it. In fields like robotics and Reinforcement Learning (RL), agents can also access information about the environment through multiple sensors; yet redundancy and complementarity between sensors is difficult to exploit as a source of robustness (e.g.\ against sensor failure) or generalization (e.g.\ transfer across domains). Prior research demonstrated that a robust and flexible multimodal representation can be efficiently constructed based on the cognitive science notion of a `Global Workspace': a unique representation trained to combine information across modalities, and to broadcast its signal back to each modality. Here, we explore whether such a brain-inspired multimodal representation could be advantageous for RL agents. First, we train a `Global Workspace' to exploit information collected about the environment via two input modalities (a visual input, or an attribute vector representing the state of the agent and/or its environment). Then, we train a RL agent policy using this frozen Global Workspace. In two distinct environments and tasks, our results reveal the model's ability to perform zero-shot cross-modal transfer between input modalities, i.e.\ to apply to image inputs a policy previously trained on attribute vectors (and vice-versa), without additional training or fine-tuning. Variants and ablations of the full Global Workspace (including a CLIP-like multimodal representation trained via contrastive learning) did not display the same generalization abilities.
\end{abstract}

\section{Introduction}
\label{sec:introduction}

Humans gather information from the world through multiple sources, leading to a rich and robust representation of their environment. Similarly, non-human agents should also learn to establish meaningful connections between information from different modalities.
Such multimodal representation learning offers distinct advantages for decision-making and in particular in Reinforcement Learning.
The benefits are evident when considering scenarios where one sensory input is noisy or unavailable. For instance, humans will be able to navigate in a room with subdued lighting where vision is compromised, as they can rely on other senses (hearing, touch...) to gather information about their environment.
In decision-making the ability to establish links between modalities allows more efficient problem-solving, because information from one sense can be leveraged to complete or verify data from another.

For these reasons, it seems advantageous to take inspiration from human multimodal integration and apply this to embodied RL agents, e.g. for robotics. A popular theory in cognitive science about how the brain handles multimodal information is the `Global Workspace Theory' \citep{baars_cognitive_1988,dehaene_neuronal_1998}. According to this theory, different specialized modules compete to encode their information into a shared space called the Global Workspace. The shared representation is then broadcast back to all modules, leading to a unified interpretation of the environment. According to the theory, this last step corresponds to our inner experience. Importantly, compared to the unimodal representations in each specialized module, the shared representation enables multimodal \emph{grounding}~\citep{silberer_grounded_2012, kiela_multi-_2015, pham_found_2019}, by linking objects and their properties across modalities. A deep learning-compatible adaptation of this theory has been proposed by \cite{vanrullen_deep_2021}. The suggested model must meet several criteria (Fig~\ref{fig:GW}): an alignment of the different latent representations and the capacity to translate from one modality to the other and to broadcast signals from the Global Workspace back to each module; ideally, the model can be trained in a semi-supervised setting with unsupervised cycle-consistency objectives. An initial implementation was reported in \cite{devillers_semi-supervised_2023}, and shown to provide reliable multimodal representations that could be leveraged advantageously for downstream classification tasks, all with minimal supervision. 

In this work, we explore the use of a similar multimodal representation, inspired by the Global Workspace Theory, in the context of RL tasks. In particular, we show that this model is capable of zero-shot cross-modal policy transfer, in two different environments (see section~\ref{sec:env}), each with two modalities (vision: RGB images; attributes: a vector description of the agent and its environment). The first environment is called \textit{Factory}, a virtual factory environment simulated in \cite{webots_httpwwwcyberboticscom_nodate}; the second one is called \textit{Simple Shapes} and made of simple geometric shapes. We chose attributes and RGB images as our two modalities because they share common information without completely overlapping, particularly in the \textit{Factory} environment (see section~\ref{sec:env}). 
Each modality must independently provide enough information to inform a \textit{unimodal} policy, and subsequently allow us to measure the potential advantages of a \textit{multimodal} representation (such as a Global Workspace).




\section{Related Work}
\label{sec:biblio}

Representation learning for Reinforcement Learning is a vast and evolving field. \citet{712192} already discussed the importance of compact representations for an RL agent. Deep Generative models, such as Variational Autoencoders (VAEs), have the capability to encode raw data into a compact and disentangled latent space. Pioneering work by \cite{watter_embed_2015} and \cite{finn_deep_2016} used this approach to encode representations for Reinforcement Learning, enhancing learning efficiency from high-resolution images. Compact representations are also crucial for algorithms relying on a World Model, such as the one introduced by \cite{ha_world_2018}.
Further studies \citep{wang_disentangled_2023,friede_learning_2023,higgins_darla_2017} showed that learning disentangled environmental representations from a VAE enables agents to develop policies robust to some shifts in the original domain. Additionally, encoding observations in a well-structured space can be achieved through contrastive learning~\citep{laskin_curl_2020,du_curious_2021}. With this method, \cite{gupta_learning_2017} were even able to measure policy transfer between robots having different numbers of joints.

Representation learning has now extended to multimodal RL setups. \cite{lee_making_2019} use fusion mechanisms with Deep Neural Networks to handle multiple sources of observations. \cite{singh_scene_2023} align visual latent representations with graphs using a contrastive loss, while \cite{hafner_mastering_2023} extend the work of \cite{ha_world_2018} by using concatenated multimodal inputs for a world model.
In a similar vein, \cite{silva_playing_2020} extend DARLA's work~\citep{higgins_darla_2017} to two modalities: sound and vision. They employ a multimodal VAE \citep{yin_associate_2017} and align representations through an additional KL loss between the two modality-specific latent spaces. This AVAE model, like ours, allows zero-shot cross-modal policy transfer, e.g. training the policy with visual inputs and using audio inputs during inference. Thus, we will use this model as a baseline to compare against our approach.

Other multimodal representation learning models like CLIP~\citep{radford_learning_2021} have been proposed to align two (or more) latent representations, and therefore to create a common space that can be used for downstream tasks. However such models require very large amounts of paired data between modalities to learn the aligned representation in a supervised way; in a robotic context, such paired data can be difficult to obtain. In addition, it has been shown that the contrastive alignment objective of CLIP tends to discard potentially important modality-specific information \citep{devillers_does_2021}. In our study, these two factors are investigated through ablation studies. First, we remove cycle-consistency objectives and train the model in a fully-supervised way. Second, we also remove the broadcast property (the ability to project global-workspace information back to each specialized module), leading to a contrastive-alignment version of our model similar to CLIP. As will be described below, both manipulations severely impair our model's ability to transfer policies between modalities. 

\section{Problem Formulation}
\label{sec:pb_formula}

\begin{figure}[ht]
    \begin{center}
        \includegraphics[scale=0.18]{Figures/All_envs_2.pdf}
    \end{center}
    \caption{A: Overview of the general approach. Raw attributes are encoded in their latent representation thanks to pre-trained models (VAE for images and Normalization for attributes). Latent image or attribute representations can be encoded into a shared space $z \in \mathcal{Z}$ (the Global Workspace or GW) via encoders $e_v$ and $e_{attr}$ (respectively). The policy is trained (solid arrows) with observations from a given modality (here vision), with GW frozen. At inference time the policy can be tested with observations from a different modality (here attributes, dashed arrow); this is defined as \emph{zero-shot cross-modal transfer}. B: Illustration of the two environments and tasks: Factory (left) and Simple Shapes (right). Example images and attributes are presented for each. For \textit{Factory}, the agent must reach the table by rotating and moving forward or backward. For \textit{Simple Shapes}, the agent must place the object at the center and pointing upwards, by moving to the right, left, top or down and rotating. }
    \label{fig:presentation}
\end{figure}

Let $\mathcal{E}$ represent an environment, whose state at time t leads to an observation $o_t \in \mathcal{O}$, described as either a latent feature vector $o_t^v$ computed from an RGB image, or an attribute vector $o_t^{attr}$. Based on these observations, the agent executes actions $a_t \in \mathcal{A}$ and receives a resulting reward $r_{t+1}$.

In this study, we first train a model to learn a representation $z_t \in \mathcal{Z}$ with two encoders $z_t^{attr} = e_{attr}(o_t^{attr})$ and $z_t^v = e_v(o_t^v)$. This step follows the approach previously described by~\cite{devillers_semi-supervised_2023}, leading to a shared representation across modalities, i.e. a Global Workspace (GW). In a second step, with GW frozen, a policy $\pi$ is trained to map GW-encoded observations from a specific training source $o\in\mathcal{O}^{train}$, with $train \in \{attr, v\}$, to actions $a \in \mathcal{A}$. During inference, the policy can potentially be transferred to another observation source $\mathcal{O}^{test}$, where $test\in \{attr,v\}, test \neq train$. The process is illustrated in Figure~\ref{fig:presentation}A, and the two training steps are further detailed below.

\subsection{GW for multimodal Representation Learning}
\label{sec:representation}

\begin{figure}[ht]
    \begin{center}
        \includegraphics[scale=0.6]{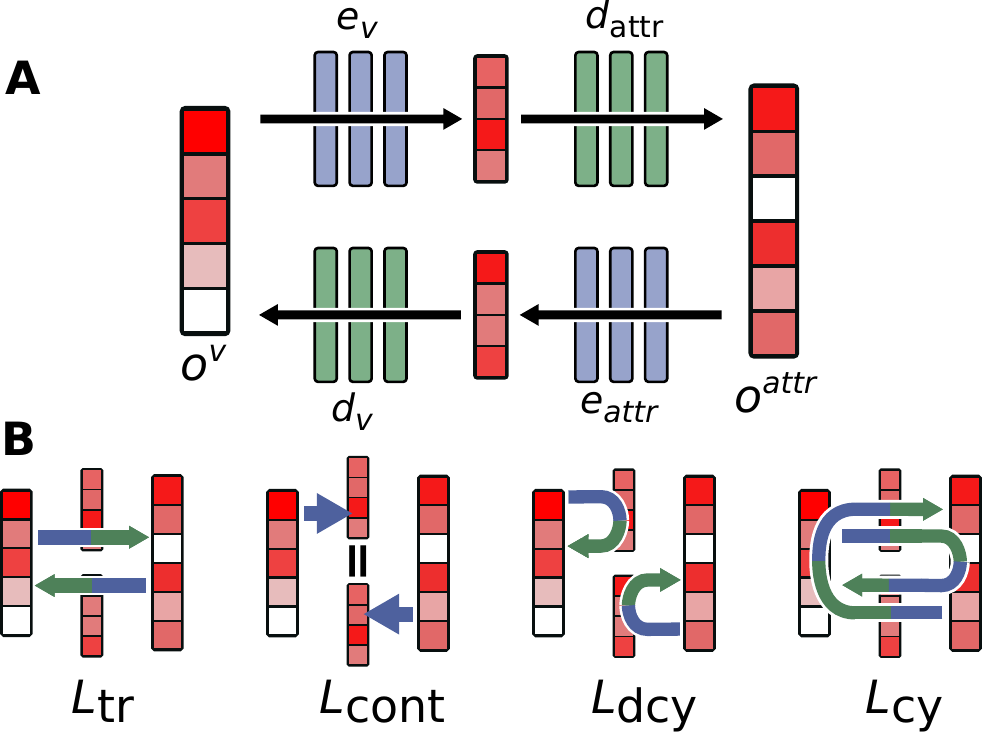}
    \end{center}
    \caption{A: A generic view of the architecture of the Global Workspace where $o^v$ and $o^{attr}$ are encoded representations of the two modalities (vision, attributes). $e_v$, $e_{attr}$ are feed-forward encoders into the GW representation, and $d_v$, $d_{attr}$ are feed-forward decoders. Encoded representations of the two modalities $e_v(o^v)$ and $e_{attr}(o^{attr})$ are separate in the architecture, but can be aligned by virtue of the training objectives (illustrated in B), resulting in a shared GW. B: Illustration of the losses used during training of the encoders and decoders. The arrows represent the path used by the data to compute the losses. $L_{tr}$ and $L_{cont}$ are supervised losses for translation and contrastive alignment, respectively; they require paired training samples across the two modalities. In contrast, $L_{dcy}$ and $L_{cy}$ are self-supervised losses for demi-cycle and full-cycle consistency, respectively; they can be trained with unpaired samples from each modality.}
    \label{fig:GW}
\end{figure}

We closely follow the training setup described in~\cite{devillers_semi-supervised_2023}. That study evaluated the properties of a multimodal GW for low-resource semi-supervised training, and for downstream classification tasks; here, we are interested in applying such a system to train an RL agent. As in this previous study, we consider a setting where matched training data across modalities can be scarce or difficult to obtain, yet we have access to potentially large amounts of unimodal data (without matching labels in the other modality). Thus, we sample unimodal observations from two sets $\mathcal{U}_{attr}$ and $\mathcal{U}_v$, and paired multimodal observations from the subset $\mathcal{M}=\mathcal{U}_{attr} \cap \mathcal{U}_v$, composed of observations that are paired across both unimodal sets. The training datasets $\mathcal{U}_v$ and $\mathcal{U}_{attr}$ are collected by uniformly sampling the environment in \textit{Simple Shapes}; for \textit{Factory} we sampled with a constraint that the table should be at least partially visible from the robot's viewpoint.

As proposed in~\cite{vanrullen_deep_2021,devillers_semi-supervised_2023}, we do not use raw images or attributes as inputs to the GW, but encoded representations into a unimodal latent space. For images, we use a Variational Autoencoder (VAE), pretrained using the set $\mathcal{U}_v$ (see Appendix \ref{sec:app_models} and \ref{sec:VAE_exp} for details) ; for attributes, we simply normalize them between -1 and 1. Then, we train the GW itself, composed of a set of encoders for each modality $\{e_v, e_{attr}\}$ with their corresponding decoders $\{d_v, d_{attr}\}$ (Figure~\ref{fig:GW}A). The role of the encoders is to project the two unimodal latent representations onto a shared one (the GW), where they should be aligned. The role of the decoders is to allow broadcast from GW back to the unimodal representations.

To train the network, four different losses are used ~\citep{devillers_semi-supervised_2023} (see Supplementary Material for losses definitions). The translation ($L_{tr}$) and contrastive alignment ($L_{cont}$) losses are supervised losses, optimized using the set $\mathcal{M}$. The full-cycle ($L_{cy}$) and demi-cycle ($L_{dcy}$) consistency losses are optimized using the full sets $\mathcal{U}_{attr}$ and $\mathcal{U}_v$. Figure~\ref{fig:GW}B illustrates how these losses are computed using the encoders and decoders of the GW. The total loss is a weighted sum of these four. \cite{devillers_semi-supervised_2023} described implicit relations between the different losses, such that optimizing a subset of the losses can indirectly improve the others. By combining the four losses, the GW model optimizes the desired criteria of multimodal representation alignment and broadcast, while taking full advantage of unsupervised training data. 

\subsection{Policy Learning and cross-modal transfer}
\label{sec:policy}

We use Proximal Policy Optimization (PPO), a widely adopted Reinforcement Learning algorithm introduced by~\cite{schulman_proximal_2017}. We also tested Advantage Actor Critic (A2C) introduced by~\cite{mnih_asynchronous_2016}, to validate our results on another algorithm (see Supplementary Materials). These two algorithms were implemented with the Stable-baselines3 library \citep{stable-baselines3}.

To obtain an upper baseline for cross-modal transfer, we train two policies in a more classical way using only unimodal information (the two policies' inputs are the unimodal representations of images $o^v$ or attributes $o^{attr}$). This is compared with policies trained from GW-encoded representations of the observations, and tested either with observations from the same modality or from the opposite modality (i.e. zero-shot cross-modal transfer). 

While our main test relies on a GW trained using all four losses (Figure~\ref{fig:GW}B), we also trained policies from GW models optimized with fewer losses, serving as ablations of the full model. A GW trained in a fully supervised way (without the cycles losses $L_{cy}$ and $L_{dcy}$) serves to assess the impact of semi-supervision, especially in low-data regimes (i.e., with few paired data in $\mathcal{M}$). We also trained a policy using a GW trained only with a contrastive loss $L_{cont}$. This ablation evaluates the impact of ``broadcast'' on the performance, and serves as a CLIP-like baseline because it is trained with the same alignment objective as CLIP~\citep{radford_learning_2021}. Finally, we compare our GW to an adaptation of the AVAE model used in~\cite{silva_playing_2020}. We modify their visual VAE to match the architecture of our own visual VAE in each environment; we also replace their audio VAE by an attribute VAE, with an architecture adapted to match the dimensions of our attribute vectors (see Supplementary Material for architecture details). This transition from audio to attribute VAE also leads to a change in the reconstruction loss: we use the same attribute reconstruction loss as the one used in the GW (see Supplementary Material). Apart from these architectural changes, the AVAE model is trained in a supervised way (on the paired multimodal set $\mathcal{M}$), as described in the original paper ~\citep{silva_playing_2020}.

For both environments, we evaluate policies based on multimodal systems (GW, GW without cycles, CLIP-like, AVAE) trained with two data regimes: either a large amount of matched data (500 000 samples for \textit{Simple Shapes} and 200 000 for \textit{Factory}), $\mathcal{M} \equiv \mathcal{U}_{attr} \equiv \mathcal{U}_v$ (full data regime), or a small amount of paired data (low data regime: $\mathcal{M}$ contains 1/4th of the full dataset for \textit{Factory}, 1/100th of the full dataset for \textit{Simple Shapes}). This assesses the impact of the unsupervised cycle-consistency losses, and the performance of fully supervised models in a low data regime.





\section{Environments}
\label{sec:env}

We evaluate our approach on two different environments. Each one captures observations across the same two modalities: attributes describing the state of the agent, or an RGB image. The first environment, called `Factory' is a simulated factory shop floor in a robotic simulator: Webots. The second environment, named `Simple Shapes' because it depicts a 2D shape on a dark background, is simulated directly using a Python-based Gymnasium environment~\citep{towers_gymnasium_2023}.

\subsection{Factory Environment}

Simulated in Webots, this environment represents a factory-like shop floor with a Tiago robot and a table. The agent receives RGB images (128x128 pixels) from the robot's viewpoint, or a set of seven attributes describing the robot and table states (Figure~\ref{fig:presentation}B). Robot state attributes include position $(x_r, y_r)$ and rotation $\theta_r$. Table state attributes include position $(x_t, y_t)$, rotation $\theta_t$, and color $h_t$. The color is defined only by the Hue of HSV, with saturation and value set to $1$ to retain high-contrast colors. The final attribute state vector concatenates attribute transformations: applying cosine and sine to angles, normalizing all attributes between -1 and 1, and decomposing the table's Hue into a cosine-sine vector.

This environment displays an asymmetry between modalities, whereby images only provide partial information while attribute vectors offer exact information, even when the robot is not facing the table. At the beginning of each episode, table attributes are randomly sampled within their domains. The robot is placed near the center with a random angle, and the agent's goal is for the robot to reach the table. The agent directly controls the position and rotation of the robot. The robot can move forward/backward and rotate (by a maximum of 5cm and $\frac{\pi}{16}$ radians during each step). Collisions with simulation objects (e.g. walls) lead to episode termination with a penalty of $-10000$. At each timestep, the reward is equal to minus the distance between the robot and the table (in meters) minus $10\times$ the angle (in radians) between the robot orientation and the robot-table vector, thus penalizing the agent for not facing the table. This approach aims to guide the robot to first locate the table by rotating and then move towards it, dividing the learning into two distinct goals and enhancing performance in scenarios where the agent relies solely on the robot's vision. When the robot reaches the table, the episode concludes with no additional reward. 

\subsection{Simple Shapes Environment}

The second environment, called `Simple Shapes', was introduced in~\cite{devillers_semi-supervised_2023}. The agent can receive two types of observations: $32\times32$ pixel RGB images of a 2D shape on a black background, or a set of eight attributes directly describing the environment's state (Figure~\ref{fig:presentation}B). There are three different types of shapes, an egg-like shape, an isosceles triangle, and a diamond. They are represented by the variable $shape \in \{0,1,2\}$. The shape possesses a size $s \in [s_{min}, s_{max}]$, a position $(x,y) \in [\frac{s_{max}}{2}, 32-\frac{s_{max}}{2}[^2$, a rotation $\theta \in [0, 2\pi[$ and an HSL color $(c_h,c_s,c_l) \in [0,1]^2 \times [l_{min},1]$. The final attribute state vector concatenates transformations of these attributes: decomposing the rotation angle $\theta$ into $(c_{\theta}, s_{\theta}) = (cos(\theta), sin(\theta))$; translating HSL colors to the RGB domain, expressing the $shape$ variable as a one-hot vector of size three, and normalizing all the variables between -1 and 1.

At the beginning of each episode, attributes are randomly sampled within their respective domains. The agent's goal is to move the shape to the center of the image at $(x,y)=(16,16)$ and align it to point to the top, $\theta=0$. Actions available to the agent include moving the shape by one pixel in cardinal directions (left, right, up, or down) and rotating the shape by an angle of $\frac{\pi}{32}$ clockwise or anti-clockwise. The reward is initialized at zero. At each timestep, the reward is equal to minus the current distance (in pixels) between the shape's position and the image center minus $10\times$ the smallest angle (in radians) between the shape's orientation and the null angle. The episode ends when the shape reaches the goal state, with no additional reward. 

\section{Results}
\label{sec:res}

\begin{figure}[ht!]
    \begin{center}
        \includegraphics[width=\linewidth]{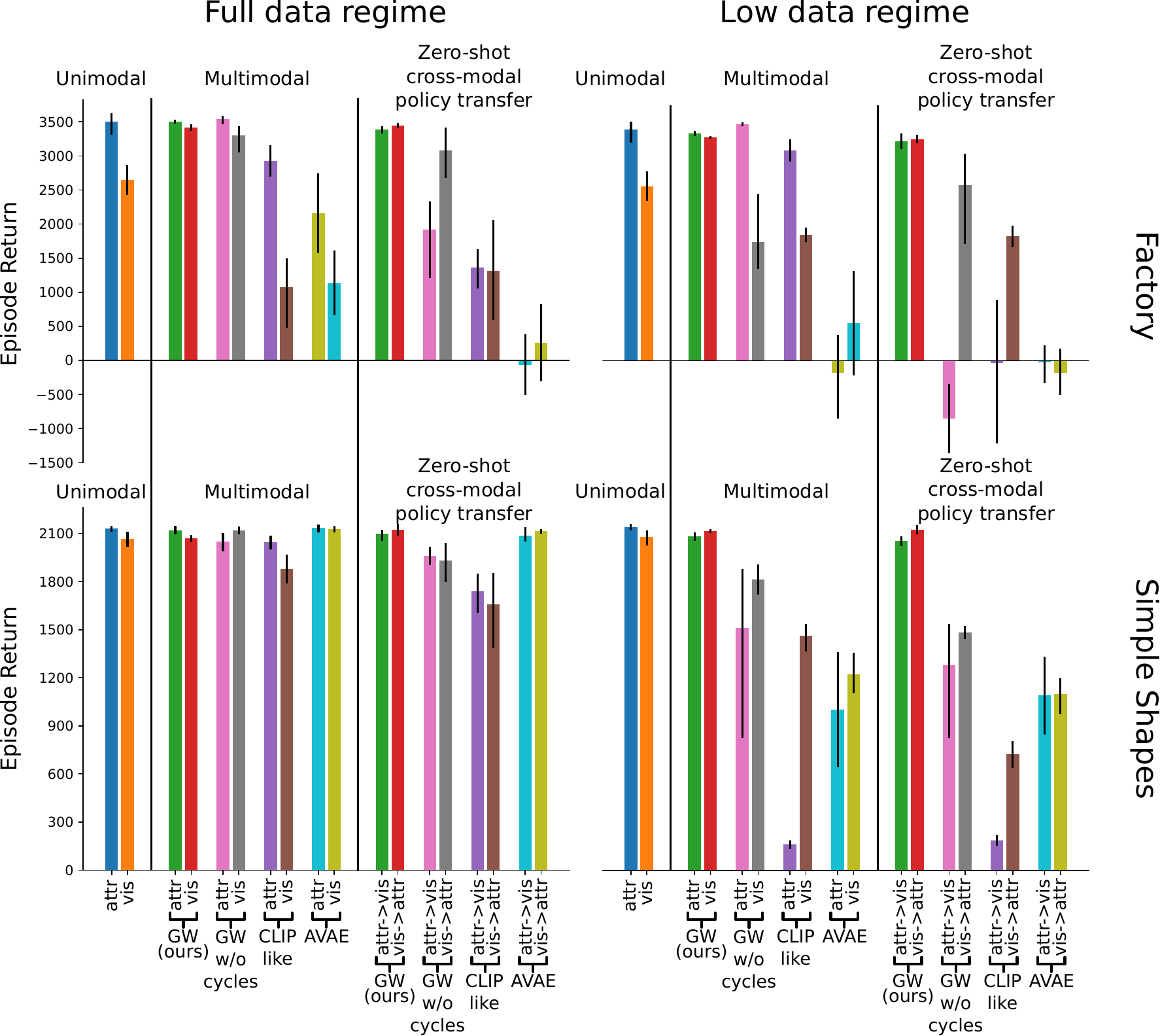}
    \end{center}
    \caption{Performance (average episode return) of PPO trained using different latent representations and tested in different settings. A fixed value was subtracted from the episode return, corresponding to the performance of a fully-random policy in each environment; thus the random policy performance (chance level) is equal to zero in all plots (negative values reflect a defective strategy, e.g. systematically hitting walls and receiving penalties). All results are averaged across five different runs (different random seeds for policy training), and the error bars reflect 95\% confidence intervals computed via bootstrapping. Models trained in the \textit{Factory} environment are plotted in the top row, and in the bottom row for the \textit{Simple Shapes} environment. Multimodal networks trained with all paired data are plotted in the left column (Full data regime); in the right column, the networks only have access to a subset of multimodal paired data (Low data regime). Each plot is divided into three parts: PPO trained directly from a \emph{unimodal} latent representation; PPO trained and tested on the same \emph{multimodal} latent representation; PPO trained on one multimodal latent representation and tested on the other (\emph{zero-shot cross modal policy transfer}). In any given plot, bars sharing the same color depict the same trained model, tested in different settings.}
    \label{fig:PPO_res}
\end{figure}

The performance (average episode return) of policies trained (via the PPO algorithm) using latent representations from different models (GW and its baselines) and in different test settings is shown in Figure~\ref{fig:PPO_res}. Results for the \textit{Factory} environment are shown in the top panels, and in the bottom panels for the \textit{Simple Shapes} environment. In each case, models trained with a Full data regime are plotted on the left, and with a Low data regime on the right.

We first focus on the performance of PPO trained directly on \emph{unimodal} representations, visible on the left part of each plot in Figure~\ref{fig:PPO_res}. As expected, \emph{unimodal} PPO acts as an upper baseline in the \textit{Simple Shapes} environment, which is fully observable from each input modality. This is not the case in \textit{Factory}, where PPO trained from attributes performs better than from vision; this highlights the asymmetry between visual inputs (partial observation) and attributes (entire state observation) in this environment.

The performance of PPO trained and tested on \emph{multimodal} latent representations obtained in a Full data regime are reported in the middle part of the two left plots in Figure~\ref{fig:PPO_res}. In both environments, GW and GW without cycles yield similar rewards as the upper baseline (PPO trained directly from attributes). AVAE achieves similar performance in \textit{Simple Shapes}, but degraded performance in \textit{Factory}. Finally, the CLIP-like model performs poorly in both environments. We can also highlight that in \textit{Factory}, policies trained from GW and (to some extent) GW w/o cycles are able to bridge the performance asymmetry between vision and attribute inputs. This is an example of \emph{multimodal grounding} in the GW, whereby the learned multimodal latent representation of visual inputs is richer and more informative for a downstream decision task than the unimodal visual latent representation. The difference with results from the CLIP-like model reveals the importance of adding broadcast objectives in addition to contrastive alignment.

In the Full data regime scenario, both supervised and semi-supervised GW models (with and without cycles) perform near-optimally when trained and tested on the same \emph{multimodal} latent representations. However, the GW cycles are particularly important when we consider the Low data regime scenario (middle part of the plots on the right in Figure~\ref{fig:PPO_res}). Here, we actually observe a drop in PPO performance for all the models in at least one input modality, except for the full GW. The decreased performance of GW w/o cycles highlights the crucial role played by the unsupervised cycle-consistency objectives in maintaining broadcast and alignment properties when the amount of multimodal paired data is low.

Finally, the zero-shot cross-modal policy transfer capabilities are shown on the right part of each plot of Figure~\ref{fig:PPO_res}. In both the full and low data regimes, and for both environments, the full GW allows for nearly optimal zero-shot transfer between modalities: a policy trained and tested on GW latent representations of attributes performs equally well when tested on GW latent representations of images (green bars), and vice-versa (red bars). The AVAE model is the only other model that permits a similar zero-shot transfer, but only in one of the four experimental settings---\textit{Simple Shapes} in the Full data regime. In the Low data regime of \textit{Simple Shapes} and in both regimes of \textit{Factory}, the policy trained in one AVAE modality does not transfer well to the other. This is also the case for the CLIP-like baseline and for the GW w/o cycles ablation, in all four experimental settings.

In summary, policies learned from a GW latent representation are particularly efficient, and in some cases (e.g., \textit{Factory}) can even surpass policies trained from unimodal representations. In addition, only policies trained from GW latents could systematically generalize to the opposite modality (zero-shot cross-modal transfer). We found that relying only on a contrastive alignment objective to establish a multimodal space (like CLIP) was insufficient. The introduction of broadcast objectives (supported by the GW decoders, see Figure~\ref{fig:GW}) compels the GW encoders to retain most information present in the original unimodal latents, so that they can be accurately reconstructed by the broadcast operation. Such a GW can be trained in a purely supervised way (GW w/o cycles) when both modalities provide fully-observable information (\textit{Simple Shapes}) and when large amounts of paired multimodal data are available for supervised training (Full data regime). In all other scenarios, the inclusion of unsupervised cycle-consistency objectives (full GW model) proves beneficial in preserving information and maintaining alignment between multimodal representations.

\section{Conclusion}
\label{sec:conclusion}

Our study applied a multimodal representation learning approach previously proposed by~\cite{devillers_semi-supervised_2023} (an adaptation of the Global Workspace Theory from Cognitive Science) to the training of an RL agent. The implemented model enables the construction of a multimodal latent space, allowing the encoding of unimodal information and exploiting the synergies between the different modalities. 
We demonstrated the capability of a GW to enable zero-shot cross-modal policy transfer, illustrating the adaptability and generalization of the learned policies across diverse modalities. Additionally, we highlighted the potential advantages of employing a semi-supervised learning framework, as seen in GW with cycle-consistency, especially in scenarios where data collection can be costly. Using a GW to generate multimodal representations, instead of other existing methods such as CLIP~\citep{radford_learning_2021} or AVAE~\citep{silva_playing_2020}, was found to improve policy performance as well as zero-shot policy transfer across modalities.
This approach not only showcases the potential of the Global Workspace Theory in enhancing the performance of RL agents, but also opens avenues for the development of more robust and versatile artificial intelligence systems capable of seamlessly transferring knowledge between different sensory domains. 
One important step towards generalizing our findings to real-world environments will be to test other modalities than vision and attributes, such as textual descriptions or proprioception (joint positions of the robot). Using sentences instead of attributes to describe the agent's state may not have a strong impact in our very controlled environments, but it could present a more significant challenge in real-world settings.



\subsubsection*{Acknowledgments}
\label{sec:ack}
This work was supported by an ANITI Chair (ANR grant ANR-19-PI3A-004), an ANR grant COCOBOT (ANR21-FAI2-0005) and by ``Défi Clé Robotique centrée sur l'humain'' funded by Région Occitanie, France. This research is also part of a project that has received funding from the European Research Council (ERC) under the European Union’s Horizon 2020 research and innovation programme (Grant agreement No.101096017). It was performed using HPC resources from CALMIP (Grant 2020-p20032).

\bibliography{main}
\bibliographystyle{rlc}

\newpage
\appendix

\section{Model Parameters}
\label{sec:app_models}

In this Appendix, we provide details about our models' implementation, starting with the $\beta$-VAE used in both visual environments: \textit{Simple Shapes} (Table~\ref{table:VAE_SS}) and \textit{Factory} (Table~\ref{table:VAE_Fac}). In the VAE encoder, all convolutions have a padding of 1, a stride of 2, and a kernel-size of 4. For the decoders, in \text{Simple Shapes} (Table~\ref{table:VAE_SS}), the transposed convolutions have a padding of 1, a stride of 2, and a kernel size of 4, except the first one which has a stride of 1. The final convolution has a stride of 1 and a kernel size of 4.
In \text{Factory} (Table~\ref{table:VAE_Fac}), the transposed convolutions have a padding of 2, a stride of 2, and a kernel size of 5, except the first one which has a stride of 1 and a kernel size of 8 without padding. The final convolution has a stride of 1 and a kernel size of 5. For both environments the $\beta$ value was set to 0.1. The $\beta$-VAE was always trained with the entire set $\mathcal{U}_v$ in both environments (500 000 images in \textit{Simple Shapes} and 200 000 images in \textit{Factory})
\begin{table}[h!]
    \begin{center}
        \begin{tabular}{l|l}
    VAE encoder ($2.8M$ params) & VAE decoder ($3M$ params)\\
    \hline
    $x\in \mathbb{R}^{3\times 32\times 32}$ & $z\in \mathbb{R}^{12}$ \\
    $\text{Conv}_{128} - \text{BN} - \text{ReLU}$ & $\text{FC}_{8\times8\times 1024}$ \\
    $\text{Conv}_{256} - \text{BN} - \text{ReLU}$ & $\text{ConvT}_{512}-\text{BN}-\text{ReLU}$ \\
    $\text{Conv}_{512} - \text{BN} - \text{ReLU}$ & $\text{ConvT}_{256}-\text{BN}-\text{ReLU}$ \\
    $\text{Conv}_{1024} - \text{BN} - \text{ReLU}$ & $\text{ConvT}_{128}-\text{BN}-\text{ReLU}$ \\
    $\text{Flatten} - \text{FC}_{2\times 12}$ & $\text{Conv}_{1}-\text{Sigmoid}$ \\
        \end{tabular}
    \end{center}
    \caption{Architecture and number of parameters of the VAE used in the \textit{Simple Shapes} environment.}
    \label{table:VAE_SS}
\end{table}

\begin{table}[h!]
    \begin{center}
        \begin{tabular}{l|l}
    VAE encoder ($2.8M$ params) & VAE decoder ($5M$ params) \\
    \hline
    $x\in \mathbb{R}^{3\times 128\times 128}$ & $z\in \mathbb{R}^{10}$ \\
    $\text{Conv}_{128} - \text{BN} - \text{ReLU}$ & $\text{FC}_{8\times8\times 512}$ \\
    $\text{Conv}_{256} - \text{BN} - \text{ReLU}$ & $\text{ConvT}_{256}-\text{BN}-\text{ReLU}$ \\
    $\text{Conv}_{512} - \text{BN} - \text{ReLU}$ & $\text{ConvT}_{128}-\text{BN}-\text{ReLU}$ \\
    $\text{Conv}_{1024} - \text{BN} - \text{ReLU}$ & $\text{ConvT}_{64}-\text{BN}-\text{ReLU}$ \\
    $\text{Flatten} - \text{FC}_{10}$ & $\text{Conv}_{1}-\text{Sigmoid}$ \\
        \end{tabular}
    \end{center}
    \caption{Architecture and number of parameters of the VAE used in the \textit{Factory} environment.}
    \label{table:VAE_Fac}
\end{table}

Table~\ref{table:GW_SS} and Table~\ref{table:GW_Fac} present details about the Global Workspace architecture for respectively \textit{Simple Shapes} and \textit{Factory}. The tables show the architecture for the encoder and decoder of only one modality, since they are nearly identical across modalities. Only the last Fully Connected layer of the decoders is different, outputting a vector of the original size of each domain.
\begin{table}[h!]
    \begin{center}
        \begin{tabular}{l|l}
    GW encoder ($35K$ params) & GW decoder ($50K$ params) \\
    \hline
    $\text{FC}_{128} - \text{ReLU}$ & $\text{FC}_{128} - \text{ReLU}$\\
    $\text{FC}_{128} - \text{ReLU}$ & $\text{FC}_{128} - \text{ReLU}$\\
    $\text{FC}_{128} - \text{ReLU}$ & $\text{FC}_{128} - \text{ReLU}$\\
    $\text{FC}_{}$ & $\text{FC}_{}$ \\
        \end{tabular}
    \end{center}
    \caption{Architecture and number of parameters for the encoder and decoder in the GW of one modality in \textit{Simple Shapes}}
    \label{table:GW_SS}
\end{table}

\begin{table}[h!]
    \begin{center}
        \begin{tabular}{l|l}
    GW encoder ($1.3M$ params) & GW decoder ($1.3M$ params)\\
    \hline
    $\text{FC}_{512} - \text{ReLU}$ & $\text{FC}_{512} - \text{ReLU}$ \\
    $\text{FC}_{512} - \text{ReLU}$ & $\text{FC}_{512} - \text{ReLU}$ \\
    $\text{FC}_{512} - \text{ReLU}$ & $\text{FC}_{512} - \text{ReLU}$ \\
    $\text{FC}_{512} - \text{ReLU}$ & $\text{FC}_{512} - \text{ReLU}$ \\
    $\text{FC}_{512} - \text{ReLU}$ & $\text{FC}_{512} - \text{ReLU}$ \\
    $\text{FC}$ & $\text{FC}$ \\
        \end{tabular}
    \end{center}
    \caption{Architecture and number of parameters for the encoder and decoder in the GW of one modality in \textit{Factory}}
    \label{table:GW_Fac}
\end{table}

The implementation details for AVAE are presented in Table~\ref{table:AVAE_SS} for \textit{Simple Shapes} and Table~\ref{table:AVAE_Fac} for \textit{Factory}. In both environments the parameters for the Conv and ConvT layers in the image VAE are the same as the ones used in their respective VAE in Tables~\ref{table:VAE_SS} and \ref{table:VAE_Fac}. For \textit{Simple Shapes}, the input layer of the attributes side is divided in two Fully Connected layers: one for the category of the shape (one-hot vector) and one for the rest of the attributes (continuous values).

\begin{table}[h!]
    \begin{center}
        \begin{tabular}{l|l}
    AVAE vision ($6M$ params) & AVAE attributes ($0.6M$ params)\\
    \hline
    $x\in \mathbb{R}^{3\times 32\times 32}$ & $x\in \{0,1\}^3 \times \mathbb{R}^{8}$ \\
    $\text{Conv}_{128} - \text{BN} - \text{ReLU}$ & $\text{FC}_{128} - \text{ReLU}$ \\
    $\text{Conv}_{256} - \text{BN} - \text{ReLU}$ & $\text{FC}_{128} - \text{ReLU}$ \\
    $\text{Conv}_{512} - \text{BN} - \text{ReLU}$ & $\text{FC}_{12} - \text{ReLU}$ \\
    $\text{Conv}_{1024} - \text{BN} - \text{ReLU}$ & \\
    $\text{Flatten} - \text{FC}_{2\times 12}$ & $\text{FC}_{2\times 12}$\\
    $z\in \mathbb{R}^{12}$ & $z\in \mathbb{R}^{12}$\\
    $\text{FC}_{8\times8\times 1024}$ \\
    $\text{ConvT}_{512}-\text{BN}-\text{ReLU}$ & $\text{FC}_{128} - \text{ReLU}$ \\
    $\text{ConvT}_{256}-\text{BN}-\text{ReLU}$ & $\text{FC}_{128} - \text{ReLU}$ \\
    $\text{ConvT}_{128}-\text{BN}-\text{ReLU}$ & $[\text{FC}_{3}, \text{FC}_{8} - \text{Tanh}]$\\
    $\text{Conv}_{1}-\text{Sigmoid}$ & \\
        \end{tabular}
    \end{center}
    \caption{Architecture and number of parameters of the visual and attributes VAEs of the AVAE for the \textit{Simple Shapes} environment.}
    \label{table:AVAE_SS}
\end{table}

\begin{table}[h!]
    \begin{center}
        \begin{tabular}{l|l}
    AVAE vision ($11M$ params) & AVAE attributes ($2M$ params)\\
    \hline
    $x\in \mathbb{R}^{3\times 128\times 128}$ & $x\in \mathbb{R}^{10}$ \\
    $\text{Conv}_{128} - \text{BN} - \text{ReLU}$ & $\text{FC}_{512} - \text{ReLU}$ \\
    $\text{Conv}_{256} - \text{BN} - \text{ReLU}$ & $\text{FC}_{512} - \text{ReLU}$ \\
    $\text{Conv}_{512} - \text{BN} - \text{ReLU}$ & $\text{FC}_{40} - \text{ReLU}$ \\
    $\text{Conv}_{1024} - \text{BN} - \text{ReLU}$ & \\
    $\text{Flatten} - \text{FC}_{2\times 40}$ & $\text{FC}_{2\times 40}$ \\
    $z\in \mathbb{R}^{40}$ & $z\in \mathbb{R}^{40}$\\
    $\text{FC}_{8\times8\times 1024}$ \\
    $\text{ConvT}_{512}-\text{BN}-\text{ReLU}$ & $\text{FC}_{512} - \text{ReLU}$\\
    $\text{ConvT}_{256}-\text{BN}-\text{ReLU}$ & $\text{FC}_{512} - \text{ReLU}$\\
    $\text{ConvT}_{128}-\text{BN}-\text{ReLU}$ & $\text{FC}_{10} - \text{Tanh}$\\
    $\text{Conv}_{1}-\text{Sigmoid}$ \\
        \end{tabular}
    \end{center}
    \caption{Architecture and number of parameters of the visual and attributes VAEs of the AVAE for the \textit{Factory} environment.}
    \label{table:AVAE_Fac}
\end{table}

\section{VAE exploration}
\label{sec:VAE_exp}

\begin{figure}[ht!]
    \begin{center}
        \includegraphics[height=20cm]{Figures/Fac_VAE_explore.pdf}
    \end{center}
    \caption{Latent traversal of the VAE used in \textit{Factory}. The rows represent the modified dimension and the columns the value added to the initial before decoding the latent vector.}
    \label{fig:Fac_VAE_explore}
\end{figure}

\begin{figure}[ht!]
    \begin{center}
        \includegraphics[width=\linewidth]{Figures/SS_VAE_explore.pdf}
    \end{center}
    \caption{Latent traversal of the VAE used in \textit{Simple Shapes}. The rows represent the modified dimension and the columns the values added to the initial before decoding the latent vector.}
    \label{fig:ss_VAE_explore}
\end{figure}

Figures \ref{fig:Fac_VAE_explore} and \ref{fig:ss_VAE_explore} illustrate the generation capabilities of each VAE in \textit{Factory} and \textit{Simple Shapes}. To produce these Figures an image was encoded to obtain a latent vector. Each dimension of this vector (rows) was modified by adding the value reported on top of each column, keeping the rest frozen. The modified vector was then decoded to obtain a resulting image. The image in the middle column in both Figures represents the initial image encoded in the VAE (because the change applied to the vector was null). This technique allows us to visualize the information captured by each latent dimension. The VAE from \textit{Factory} captures the robot's position and rotation well, as we can see how the background changes with different dimensions and values. For the table, the color, position and rotation are also well captured, as we can guess these information from the reconstruction. In the \textit{Simple Shapes} VAE, all attribute information is captured since Figure \ref{fig:ss_VAE_explore} shows different shapes with varying colors, positions. 

\section{GW losses details}
\label{sec:gw_loss}

As explained in~\ref{sec:pb_formula}, the Global Workspace (GW) is trained with four different losses. Here we provide details of their implementation, following~\cite{devillers_semi-supervised_2023}.

\begin{equation}
    \begin{aligned}
    &L_{tr} && = \frac{1}{2} [L_{attr}(d_{attr}(e_v(o_v^i)), o_{attr}^j) + L_v(d_v(e_{attr}(o_{attr}^j)), o_v^i)] \\
    &L_{cont} && = CONT[e_v(o_v^i), e_{attr}(o_{attr}^j)] \\
    &L_{dcy} && = \frac{1}{2} [ L_v(d_{v}(e_v(o_v^i)), o_v^i) + L_{attr}(d_{attr}(e_{attr}(o_{attr}^j)), o_{attr}^j)] \\
    &L_{cy} && = \frac{1}{2} [L_v(d_v(e_{attr}(d_{attr}(e_v(o_v^i)))), o_v^i) + L_{attr}(d_{attr}(e_v(d_v(e_{attr}(o_{attr}^j)))), o_{attr}^j)]\\
    \end{aligned}
\end{equation} 

Where $CONT()$ is the contrastive loss used in the CLIP model \citep{radford_learning_2021}. $L_{attr}$ represents the reconstruction loss used on the attributes side, which differs between the two environments. In \textit{Factory} (where all attributes have continuous values), it is computed with an MSE; in \textit{Simple Shapes} it is a combination of a negative log-likelihood for shape classes (discrete one-hot encoded values) and MSE for the other (continuous) attributes. $L_v$ represents the reconstruction loss on the visual side, computed with an MSE in both environments. The total loss is then computed as follows : 

\begin{equation} 
    \begin{aligned} 
    L_{GW} = \alpha \cdot L_{tr} + \beta \cdot L_{cont} + \gamma \cdot L_{dcy} + \theta \cdot L_{cy}
    \end{aligned}
\end{equation}

Where $\alpha, \beta, \gamma, \theta$ are hyperparameters giving more or less importance to each loss. The following table contains the hyperparameters for all Global Workspace models (and ablations) in the Full data regime in both environments.

\begin{center}
    \begin{tabular}{c|c|c|c}
& GW & GW w/o cycles & CLIP-like\\
\hline
Factory & $\alpha=1$ & $\alpha=1$ & $\alpha=0$ \\
&  $\beta=0.1$ & $\beta=0.1$ & $\beta=1$ \\
&  $\gamma=1$ & $\gamma=0$ & $\gamma=0$ \\
&  $\theta=1$ & $\theta=0$ & $\theta=0$ \\
\hline
Simple Shapes & $\alpha=1$ & $\alpha=1$ & $\alpha=0$ \\
&  $\beta=0.1$ & $\beta=0.1$ & $\beta=1$ \\
&  $\gamma=5$ & $\gamma=0$ & $\gamma=0$ \\
&  $\theta=5$ & $\theta=0$ & $\theta=0$ \\
    \end{tabular}
\end{center}

The table below shows the hyperparameters used in the Low data regime in both environments.
\begin{center}
    \begin{tabular}{c|c|c|c}
& GW & GW w/o cycles & CLIP-like\\
\hline
Factory & $\alpha=1$ & $\alpha=1$ & $\alpha=0$ \\
&  $\beta=0.1$ & $\beta=0.1$ & $\beta=1$ \\
&  $\gamma=5$ & $\gamma=0$ & $\gamma=0$ \\
&  $\theta=5$ & $\theta=0$ & $\theta=0$ \\
\hline
Simple Shapes & $\alpha=1$ & $\alpha=1$ & $\alpha=0$ \\
&  $\beta=0.1$ & $\beta=0.1$ & $\beta=1$ \\
&  $\gamma=10$ & $\gamma=0$ & $\gamma=0$ \\
&  $\theta=10$ & $\theta=0$ & $\theta=0$ \\
    \end{tabular}
\end{center}

\section{Reward details}
\label{sec:reward}

The reward in the \textit{Factory} environment is given by a combination of the distance between the robot and the table, and the angle between the orientation of the robot and the table (this is meant to encourage the policy to turn the robot facing the table, regardless of its original location):
\begin{equation}
    \label{rew:Factory}
        \begin{aligned}
          & r = - \text{distance} - 10 \times \text{angle} \\
          & r = - \sqrt{(x_r-x_t)^2 + (y_r-y_t)^2} - 10 \times | \arccos([c_{\theta_r}, s_{\theta_r}], \frac{[x_t-x_r, y_t-y_r]}{||[x_t-x_r, y_t-y_r]||_2}) |
        \end{aligned}
\end{equation}

The reward in the \textit{Simple Shapes} environment is given by a combination of the distance between the position of the shape and the center of the image, and the angle of the shape:
\begin{equation}
    \label{rew:SS}
        \begin{aligned}
          & r = - \text{distance} - 10 \times \text{angle} \\
          & r = - \sqrt{(x-16)^2 + (y-16)^2} - 10 \times | \arccos([c_{\theta}, s_{\theta}], [1,0]) |
        \end{aligned}
\end{equation}

\section{A2C in Simple Shapes scenario}
\label{sec:app_A2C}

\begin{figure}[ht]
    \begin{center}
        \includegraphics[width=\linewidth]{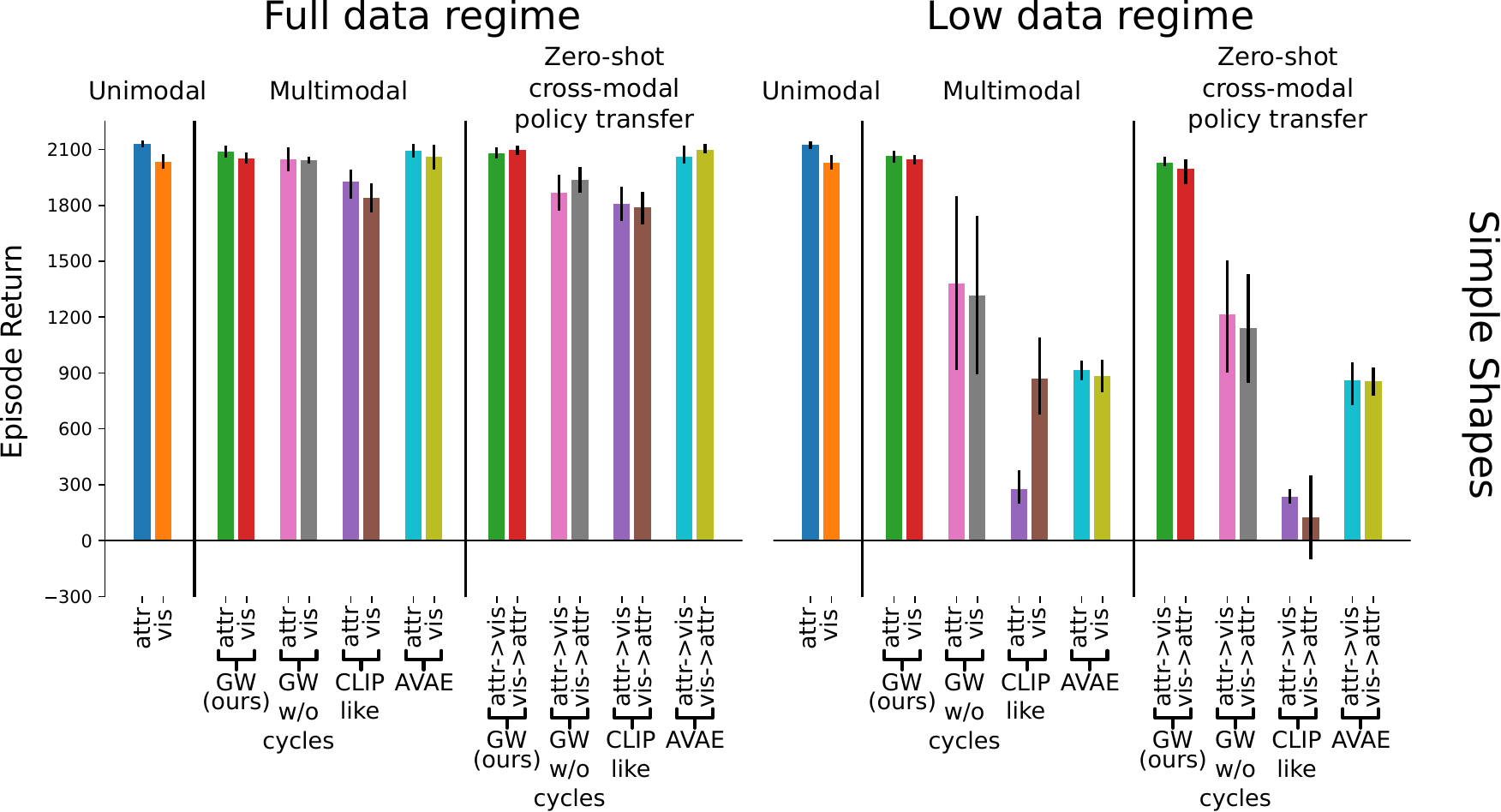}
    \end{center}
    \caption{Performance of A2C in the Simple Shapes environment. Notations and conventions as in Figure~\ref{fig:PPO_res}.}
    \label{fig:A2C_res}
\end{figure}

An additional experiment was performed in the Simple Shapes environment to verify that our results were robust to the choice of policy training algorithm. For this, we used A2C, introduced by \cite{mnih_asynchronous_2016}. Figure~\ref{fig:A2C_res} shows that the results are reproducible with this alternative algorithm (compare with Figure~\ref{fig:PPO_res}, bottom). A2C trained from a Global Workspace performs as well as when trained on unimodal representations, both in terms of absolute performance and in terms of zero-shot cross-modal transfer. AVAE performs similarly in the Full data regime, but poorly in the Low data regime. The two other models (Global Workspace without cycles and CLIP-like ablation), give worse performance in both regimes, as in the case of PPO. 

\end{document}